\documentclass[letterpaper, 10 pt, conference]{ieeeconf}

\IEEEoverridecommandlockouts
\usepackage{graphicx} 
\usepackage[table]{xcolor}
\usepackage[utf8]{inputenc}
\usepackage{amsmath,amssymb,amsfonts,cite} 
\usepackage{siunitx,textcomp} 
\usepackage[hidelinks]{hyperref} 
\usepackage{subcaption}
\let\labelindent\relax
\usepackage[inline]{enumitem} 
\usepackage[nolist]{acronym} 

\newacro{cnn}[CNN]{Convolutional Neural Network}
\newacro{gmm}[GMM]{Gaussian Mixture Model}
\newacro{gmr}[GMR]{Gaussian Mixture Regression}
\newacro{em}[EM]{Expectation Maximization}
\newcommand{\figref}[1]{\hyperref[#1]{Fig.~\ref*{#1}}}
\newcommand{\tabref}[1]{\hyperref[#1]{Table~\ref*{#1}}}
\newcommand{\secref}[1]{\hyperref[#1]{Section~\ref*{#1}}}
\newcommand{\algoref}[1]{\hyperref[#1]{Algorithm~\ref*{#1}}}

\newcommand{\subfigref}[1]{(\subref{#1})}

\newcommand{\matr}[1]{\mathbf{#1}}

\newcommand{\expfig}[2]{\begin{subfigure}[b]{.32\textwidth}
	\centering
	\includegraphics[width=\linewidth]{figures/#1.png}%
	\caption{#2}\label{fig:#1}
\end{subfigure}}

\def\bestcolor{(best viewed in color)}
\def\panda{\textit{Franka Emika Panda}}
\def\ie{\textit{i.e.,}}
\def\eg{\textit{e.g.,}}
\def\etal{\textit{et al.}}
\def\pc{point-cloud}
\def\expreg{\textit{explored regions}}
\def\unexpreg{\textit{unexplored regions}}
\def\expregsing{\textit{explored region}}

\title{\LARGE \bf
Probabilistic Surface Friction Estimation\\
Based on Visual and Haptic Measurements
}

\author{Tran~Nguyen~Le, Francesco~Verdoja, Fares~J.~Abu-Dakka, Ville~Kyrki%
\thanks{This work was supported by Academy of Finland Strategic Research Council
grant 314180 and CHIST-ERA project IPALM (326304).} \thanks{All authors are with
Intelligent Robotics Group at the Department of Electrical Engineering and
Automation, School of Electrical Engineering, Aalto University, Finland.
\texttt{\{firstname.lastname\}{@}aalto.fi}}}

\begin{document}

\maketitle
\thispagestyle{empty}
\pagestyle{empty}


\begin{abstract}
    Accurately modeling local surface properties of objects is crucial to many
    robotic applications, from grasping to material recognition. Surface
    properties like friction are however difficult to estimate, as visual
    observation of the object does not convey enough information over these
    properties. In contrast, haptic exploration is time consuming as it only
    provides information relevant to the explored parts of the object. In this
    work, we propose a joint visuo-haptic object model that enables the
    estimation of surface friction coefficient over an entire object by
    exploiting the correlation of visual and haptic information, together with a
    limited haptic exploration by a robotic arm. We demonstrate the validity of
    the proposed method by showing its ability to estimate varying friction
    coefficients on a range of real multi-material objects. Furthermore, we
    illustrate how the estimated friction coefficients can improve grasping
    success rate by guiding a grasp planner toward high friction areas.
\end{abstract}

\section{Introduction}
\label{sec:introduction}

Nowadays, robots are used extensively to perform various tasks from simple
pick-and-place to sophisticated object manipulation in complex environments from
factory floors to hospitals. For such tasks, robots are required to interact
with, and adapt to, unknown environments and objects. In order to successfully
accomplish these tasks, robots need to identify various properties of the
objects to be handled. For these reasons, identifying object models that can
represent the properties of objects has become a crucial issue in robotics.

Many object-modelling approaches have focused on object shape and geometry by
utilizing vision~\cite{autonomous_3dmodel_16,prankl_15,weise_09}. However, other
physical properties also play an important role in characterizing object
behavior during interaction and handling. In particular, surface properties such
as surface friction, texture, and roughness are vital for manipulation planning.

Various methods have been proposed to learn object surface properties from
vision~\cite{zhang2016friction,xue17,brandao16} or haptic
feedback~\cite{yoshikawa_91,cutkosky93,liu12,maria13,icra20}. Also the
combination of both vision and haptic cues~\cite{rosales14, bhat15} has been
proposed, similar to human perception~\cite{WHITAKER200859}. However, most
published works assume the surface properties to be identical across the whole
surface of the object. This assumption does not hold for many real objects,
since objects often consist of multiple materials. 

\begin{figure}
    \centering
    \includegraphics[width=\linewidth]{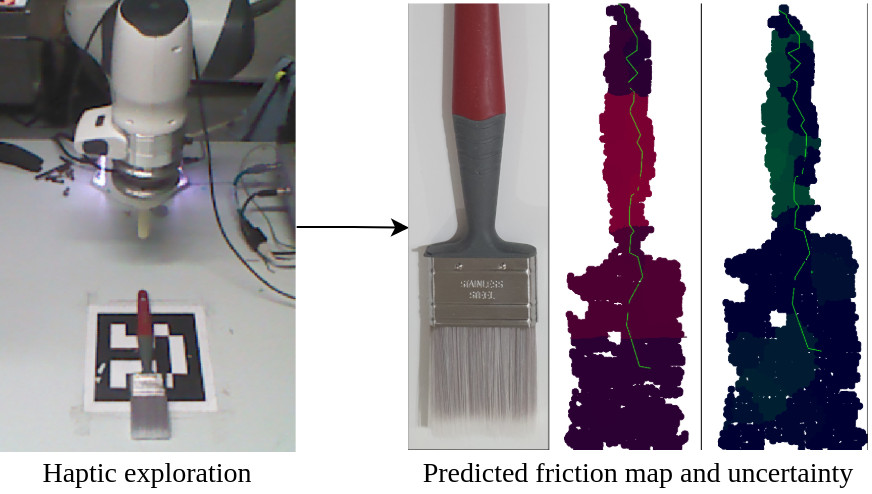}
    \caption{After the shape of an object has been captured from a camera,
    surface friction on a small part of the object is estimated using haptic
    exploration. The friction over the entire object is then predicted by
    coupling visual information with the local haptic measurements.}
    \label{fig:paper_title}
\end{figure}

To address this, we propose a method to estimate the surface properties of a
multi-material object by combining a visual prior with haptic feedback obtained
while performing haptic exploration on the object. We focus on one property, the
surface friction coefficient, but the proposed method could be applied to other
properties such as texture and roughness. The approach is based on the
assumption that visual similarity implies similarity of surface properties. By
measuring a property directly using haptic exploration over a small part of the
object, the joint distribution of visual and haptic features can be constructed.
Using the joint distribution, the measurement can then be generalized over all
parts of the object that are visible. The inference allows recovering both the
expected value of friction for each part as well as a respective measure of
prediction confidence.

The main contributions of this paper are: 
\begin{itemize}
    \item a probabilistic method to estimate the object friction coefficient
    based on visual prior and haptic feedback without being restricted by the
    assumption that objects have uniform homogeneous friction everywhere;
    \item a set of experiments on a physical robot showing the proposed method
    working on a wide range of objects, including multi-material objects;
    \item a case study, demonstrating the ability of the estimated friction
    coefficients to guide grasp planning towards areas of high friction,
    improving grasping success rate.
\end{itemize}

\section{Related work}
\label{sec:related_work}

\subsection{Friction estimation from vision}

In the context of friction estimation from vision, most recent improvements come
from the adoption of deep learning. Majority of the works focuses on recognizing
the material using images then assigning the friction coefficient based on a
material dataset. For instance, Zhang~\etal{}~\cite{zhang2016friction} presented
a material recognition method of deep reflectance code that encodes information
about the physical properties of a surface from reflectance measurements. The
predicted friction coefficient of a sample is then assigned as the average of
friction coefficients of corresponding samples in the dataset. Another approach
for material recognition was proposed by Xue~\etal{}~\cite{xue17}, where they
developed a deep neural network architecture using angular variation features.
Brandão~\etal{}~\cite{brandao16} proposed a solution to this problem by
combining a state-of-the-art deep \ac{cnn} architecture to predict broad
material classes from images with known distributions of material friction. The
predicted friction coefficient is then used to plan the locomotion of biped
robot on different terrains.

One downside of the aforementioned approaches is that wrong material recognition
will lead to wrong friction coefficient estimation. As the mentioned works
utilize different visual features to recognize the materials, the recognition
result depends heavily on the quality of visual input. However, vision is
usually impaired by occlusions, lighting condition, and location of the sensor.
Furthermore, even assuming the visual input to be perfectly gathered, these
approaches may also fail in cases such as 
\begin{enumerate*}
    \item different visual features having the same friction coefficients, or
    \item same visual features having different friction coefficients.
\end{enumerate*}
In this paper, we overcome these limitations by combining the visual input with
haptic feedback to directly estimate friction coefficients of an object. The
addition of haptic feedback offers the possibility of accessing information that
is hardly perceptible visually. In addition, the aforementioned works only
attempt to estimate surface properties for outdoor scenes and applied to mobile
robotics domain while in this work we target household objects.

\subsection{Friction estimation from haptic feedback}

The idea of using exploratory actions to estimate physical properties of objects
has been carried out in many works \cite{lederman93}. One of the earliest work
on friction estimation from haptic feedback was proposed by
Yoshikawa~\etal{}~\cite{yoshikawa_91}, where the authors described a method to
estimate the friction distribution and the centre of friction of an object by
pushing it with a mobile manipulator. Similar works on estimating friction
coefficient were carried out with different exploration actions such as
pushing~\cite{lynch93}, pressing~\cite{maeno04}, or lateral
sliding~\cite{cutkosky93,liu12,maria13,icra20}. However, all these works are
only valid under the assumption that the surface properties are identical across
the whole object surface. This limitation makes it difficult to apply the
methods on a wider range of objects including multi-material objects, something
that we address in this work. Rosales~\etal{}~\cite{rosales14} attempted to lift
this limitation by proposing a representation that consists of both shape and
friction that were gathered through an exploration strategy. They then used a
Gaussian Process to approximate the distribution of the friction coefficient
over the surface. However, the results presented in that work show that the
friction coefficient is only estimated for the regions that are explored by the
robot. Unexplored areas on the object surface are then assigned a non-sense
value. Our method, on the other hand, estimates friction coefficient also for
unexplored areas based on information gathered from explored areas.
Additionally, \cite{rosales14} only considers single-material objects such as a
paperboard box, and a metallic can for experimental evaluation. In this work, we
experimentally evaluated our method with different objects including both
single-material and multi-material objects. 
\vspace{-0.5em}

\section{Problem formulation}
\label{sec:prob_form}
\begin{figure*}
    \centering
    \includegraphics[width=\linewidth]{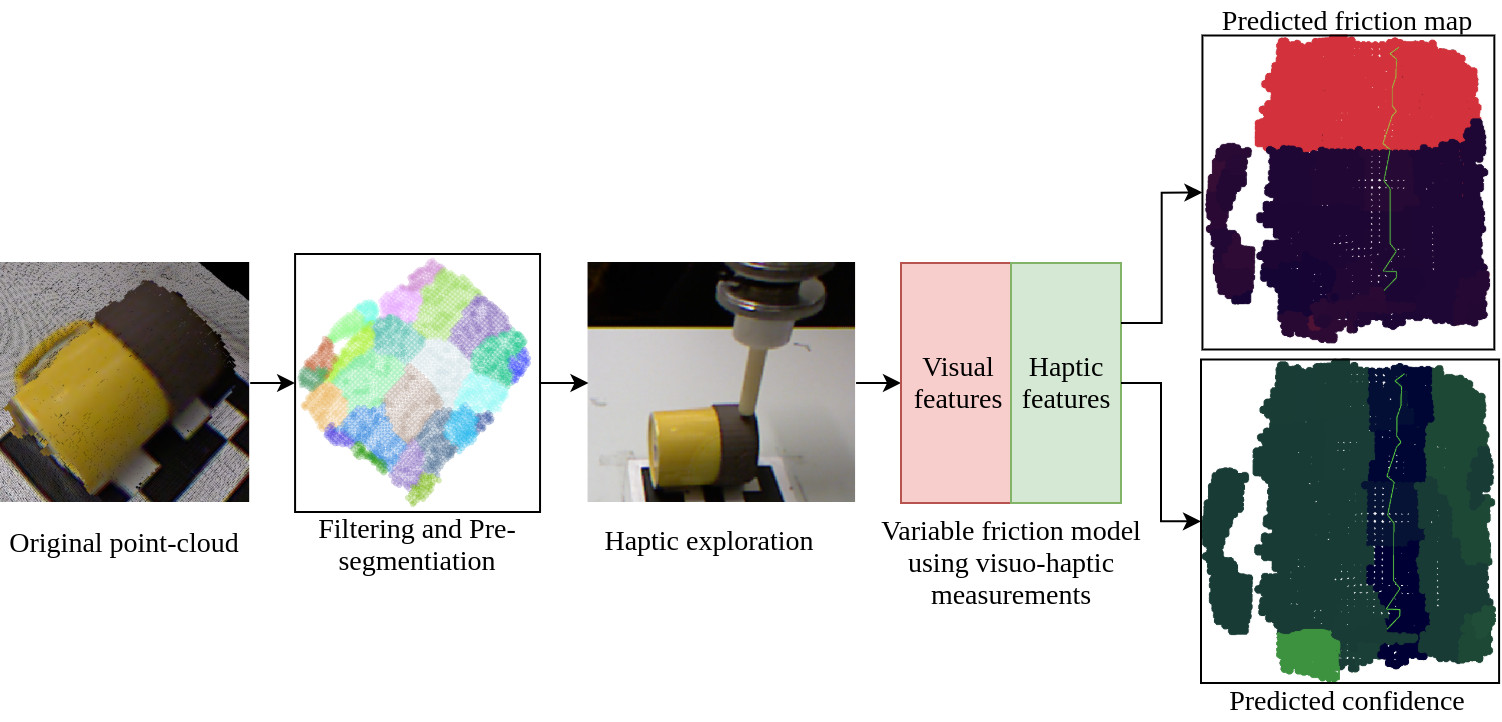}
    \caption{The proposed pipeline: the object's visual properties are first
    acquired as \pc{}, which is then filtered and pre-segmented into regions.
    The robot then performs an haptic exploration over some of the regions. The
    proposed model then is used to estimate the friction coefficient over the
    whole object, together with the corresponding confidence.}
    \label{fig:pipeline}
\end{figure*}
The problem of identifying the friction coefficients of an unknown object is
difficult. A typical approach for this problem is to utilize haptic sensory
feedback. However, haptic feedback is usually high cost, noisy, and unreliable
due to the fact that haptic sensing introduces many issues such as low
durability, low performance, high cost, and less compatibility with other
sensors~\cite{yamaguchi19}. Another way of gathering data that can be processed
for friction estimation is visual sensing. Although visual feedback is cheap and
intuitive, it does not always provide enough information to identify friction
coefficient of an unknown object. In this work, we address the problem of
estimating friction coefficient of unknown objects lying on a supporting plane
by combining both visual and haptic sensing. The goal of this paper is to
understand the correlation between the known visual and haptic feature of the
target object and, based on that correlation, to then extrapolate the unknown
haptic features from the visual features. 

Visual information about the scene is obtained by a RGB-D camera whose pose is
known relative to the robot. Haptic feedback is gathered while performing
exploratory actions on the target object using a robotic arm equipped with a
plastic finger and force/torque sensor. To make it easier for latter
representation, the regions that are touched by the robot during the haptic
exploration are called \expreg{}, and the one that are not touched \unexpreg{}.
Let $\mathbf{V}$ and $\mathbf{H}$ denote visual and haptic features,
respectively. For the \expreg{}, both $\mathbf{V}$ and $\mathbf{H}$ are known
while for \unexpreg{} only $\mathbf{V}$ is known. The goal is to infer haptic
feature $\mathbf{H}$ from the visual feature $\mathbf{V}$ of \unexpreg{} based
on the joint $\mathbf{(V,H)}$ model of the \expreg{}. In other words, the
objective is to find the conditional probability $\mathcal{P}(\mathbf{H} \mid
\mathbf{V})$ for \unexpreg{}. 

To this end, we propose to model the joint distribution $\mathcal{P}(\mathbf{V},
\mathbf{H})$ by fitting a \ac{gmm} over all the visuo-haptic features of all
points in the \expreg{}. More precisely, for an object with $n$ materials, let
us build the $C$-component \ac{gmm} that better fit the visuo-haptic data from
the \expreg{}, with one component for each of the $n$ materials, plus one
background component describing an uninformative visuo-haptic prior, \ie{} $C =
n+1$. Formally, we build a multivariate \ac{gmm} model $M$ to estimate the joint
probability distribution $\mathcal{P}(\mathbf{V},\mathbf{H})$ from gathered data
in the \expreg{}, \ie
\begin{equation} \label{eq:gmm}
    \mathcal{P}(\mathbf{V},\mathbf{H}) \sim 
    \sum_{c=1}^{C} \pi_c \mathcal{N}(\mu_c, \Sigma_c)\enspace,
\end{equation}
where $\pi_c$, $\mu_c$, and $\Sigma_c$ denote the prior probability, mean, and
covariance of the $c$-th Gaussian component respectively. Formally, let us
decompose the \ac{gmm} parameters $\mu_c$ and $\Sigma_c$ as
\begin{equation} \label{eq:mean_cov}
    \mu_c=
    \begin{bmatrix}
        \mu_c^\mathbf{V} \\ \mu_c^\mathbf{H}
    \end{bmatrix}, \quad
    \Sigma_c=
    \begin{bmatrix}
        \Sigma_c^\mathbf{V} & \Sigma_c^{\mathbf{V}\mathbf{H}} \\ 
        \Sigma_c^{\mathbf{H}\mathbf{V}}& \Sigma_c^\mathbf{H}
    \end{bmatrix}\enspace.
\end{equation}
	
After the model $M$ is fitted to a given dataset, by using \ac{gmr}
\cite{calinon2016tutorial}, we can estimate the haptic feature $\mathbf{H}_i$ at
each data point $i = 1, ..., N$ in the \unexpreg{} given input $\mathbf{V}_i$ by
means of the conditional probability $\mathcal{P}(\mathbf{H} \mid \mathbf{V})$
expressed as
\begin{equation} \label{eq:conditional}
    \mathcal{P}({\mathbf{H}_i\mid\mathbf{V}_i)} \sim
    \sum_{c=1}^{C} h_c(\mathbf{V_i})
    \mathcal{N}(\hat{\mu}_c^\mathbf{H}(\mathbf{V}_i),
    \hat{\Sigma}_c^\mathbf{H})\enspace.
\end{equation}

Then, given an input $\mathbf{V}_i$, the mean $\hat{\mu}_i^{\mathbf{H}}$ and
covariance $\hat{\Sigma}_i^{\mathbf{H}}$ of its corresponding output
$\mathbf{H}_i$ are computed by
\begin{equation} \label{eq:inference}
\begin{split}
    \hat{\mu}_{i}^\mathbf{H} =& \mathbb{E}(\mathbf{H}_i\mid\mathbf{V}_i) = 
    \sum_{c=1}^C h_c(\mathbf{V}_i)\hat{\mu}_c^{\mathbf{H}}(\mathbf{V}_i)
    \enspace,\text{ and }\\
    \hat{\Sigma}_i^\mathbf{H} =& \sum_{c=1}^{C} h_c(\mathbf{V}_i)
    (\hat{\Sigma}_c^\mathbf{H} + 
    \hat{\mu}_c^\mathbf{H}(\mathbf{V}_i)
    \hat{\mu}_c^\mathbf{H}(\mathbf{V}_i)^\intercal) \\
    & - \hat{\mu}_i^\mathbf{H}(\hat{\mu}_i^\mathbf{H})^\intercal\enspace,
\end{split}
\end{equation}
where
\begin{equation} \label{eq:inference2}
\begin{split}
    \hat{\mu}_c^{\mathbf{H}}(\mathbf{V}_i) =& \mu_c^\mathbf{H} + 
    \Sigma_c^{\mathbf{H}\mathbf{V}} (\Sigma_c^{\mathbf{V}})^{-1} 
    (\mathbf{V}_i - \mu_c^\mathbf{V})\enspace, \\[.7em]
    \hat{\Sigma}_c^\mathbf{H} =& \Sigma_c^\mathbf{H} - 
    \Sigma_c^{\mathbf{H}\mathbf{V}} (\Sigma_c^{\mathbf{V}})^{-1} 
    \Sigma_c^{\mathbf{V}\mathbf{H}}\enspace, \\[.7em]
    h_c(\mathbf{V}_i) =& \frac{\pi_c \mathcal{N}(\mathbf{V}_i\mid 
    \mu_c^\mathbf{V},\Sigma_c^\mathbf{V})}{\sum_{k=1}^{C}\pi_k 
    \mathcal{N}(\mathbf{V}_i\mid \mu_k^\mathbf{V},\Sigma_k^\mathbf{V})}\enspace.
\end{split}
\end{equation}

\section{Implementation}
\label{sec:implementation}

The system pipeline shown in \figref{fig:pipeline} consists of:
\begin{enumerate*}[label=(\roman*)]
    \item filtering and pre-segmenting the real objects,
    \item conducting an haptic exploration process to gather and couple haptic
    data with visual data, 
    \item modeling the variable friction model using visuo-haptic data, and
    \item inferring the friction coefficients and inference confidence of
    \unexpreg{} from the modelled distribution.
\end{enumerate*}

\subsection{Visual filtering and pre-segmentation}

The scene in the original \pc{} $Y$ contains a target object lying on a table.
As the pose of the camera with respect to the robot is known, we can first
remove points that are part of the supporting surface, and points that belong to
the background indicated by their distance exceeding a certain threshold. We
then obtain a filtered \pc{} $\bar{Y}$ containing only the view of the object to
be pre-segmented.

Formally, let ${y} = (\mathbf{x_p},\mathbf{V})$ denote a point in the filter
\pc{} $\bar{Y}$ ($y\in\bar{Y}$), $\mathbf{x_p}\in\mathbb{R}^3$ is the position
of the point with respect to the camera frame, and $\mathbf{V}\in\mathbb{R}^3$
is the RGB component vector representing its visual feature. It should be noted
that for objects with textured appearance, visual texture features could be used
instead of color.

As discussed in the previous section, the main goal of this work is to estimate
haptic features for \unexpreg{} based on the gathered visual and haptic features
of \expreg{}. Thus, we consider a region-based representation of the target
object. The idea is to divide the target object into a large number $N$ of
connected regions, wherein each region $i = 1, ..., N$ has its own visual
feature and haptic feature. We achieved this using a state-of-the-art supervoxel
segmentation named VCCS~\cite{vccs}. Given the filtered \pc{} $\bar{Y}$, we
defined an $N$-region segmentation as a partition $R = \{r_i\}_{i=1}^N$ of the
points of $\bar{Y}$. More precisely, the regions must satisfy the following
constrains:
\begin{equation} \label{eq:segmentation}
    \begin{aligned}
        &\forall{y}\in\bar{Y}\; 
        (\exists{r}\in R \mid y \in r)\enspace; \\
        &\forall{r}\in R \; 
        (\nexists {r'}\in R \mid r \cap r' \neq \emptyset)\enspace.
    \end{aligned}
\end{equation}

These constrains guarantees that all points in the filtered \pc{} $\bar{Y}$
belong to a region, and no point can belong to two regions. The second step in
\figref{fig:pipeline} shows the result after the filtering and pre-segmentation
step. The next step is to gather and couple haptic data with visual data through
haptic exploration. 

\subsection{Haptic exploration and tracking}

\begin{figure}
    \centering
    \includegraphics[width=0.8\linewidth]{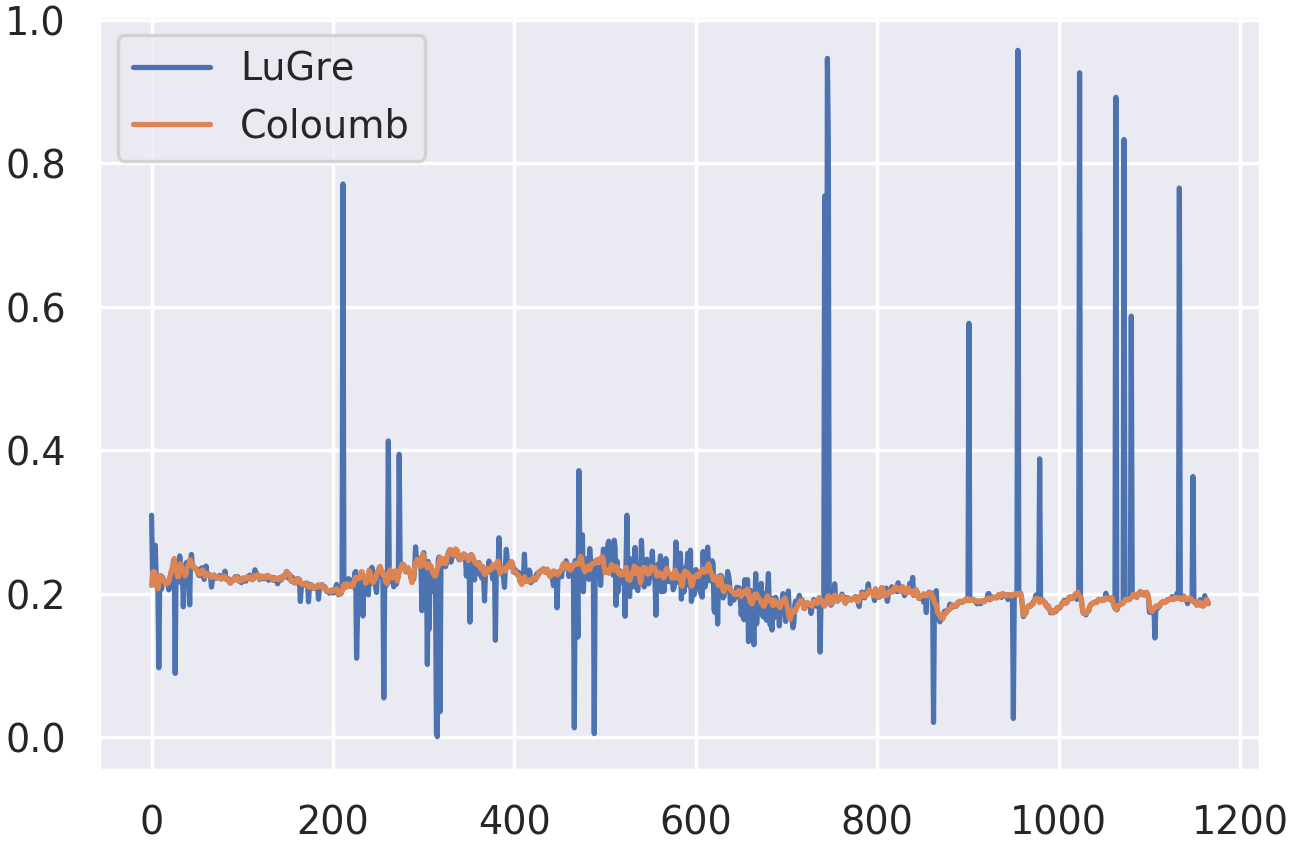}
    \caption{A qualitative comparison between friction models. The orange line
    denotes the friction coefficient estimated using Coloumb friction model,
    while the blue line denotes the one using LuGre friction model
    \bestcolor{}.}
    \label{fig:lugrevscoloumb}
    \vspace{-1.2em}
\end{figure}
In this work, haptic data is gathered using a force/torque (F/T) sensor attached
to the wrist of a robot arm. The haptic data consists of the contact point on
the object surface $\mathbf{x_c} \in \mathbb{R}^3$ with respect to the robot
base frame calculated using forward kinematics, and the contact force between
the finger and the object $\mathbf{f} \in \mathbb{R}^3$ expressed in contact
frame. The dynamic friction coefficient, which represents haptic feature
$\mathbf{H}$, can be then estimated using Coulomb friction model as
\begin{equation} \label{eq:eq_mu}
   \mathbf{H} = c.o.f. = 
   \frac{\mid{\mathbf{f}_t}\mid}{\mid{\mathbf{f}_z}\mid}\enspace,
\end{equation}
where $\mathbf{f}_t$ and $\mathbf{f}_z$ are the tangent and normal forces at the
contact point, respectively. The reason we chose Coulomb friction model is that
it produces less noisy point-wise estimation and requires less computational
time compared to others friction model such as LuGre model \cite{lugre}. This
claim is made based on the result of conducted experiments where we used both
Coloumb and LuGre friction models to estimate the friction coefficient of a
plastic plate (\figref{fig:lugrevscoloumb}) and compared the results with the
ground-truth presented in \cite{cofdatabase}. The calculated c.o.f. ranges from
0.2 to 0.35, while the ground-truth ranges from 0.2 to 0.4. In this work, we
consider dynamic friction because it is easier and more efficient to measure
compared to static friction, since static friction measurement requires starting
and stopping motion for each measurement point. Furthermore, as static friction
is often higher than dynamic friction, the latter will act as the safe lower
bound in grasping scenarios.

To obtain haptic data of an object, we perform an exploratory action called
lateral sliding on the object where we use the robot arm equipped with an F/T
sensor to slide on the object surface along a linear path. Hybrid force and
position control is used during the exploration to guarantee the contact with
the object surface. The exploration depends on the object remaining immobile. In
our case, the object is manually held to prevent the object from moving during
the exploration. While not optimal, this is a typical way to stabilize objects
under exploration procedures \cite{jamisola14}. The problem could also be solved
by using a dual arm setup to remove the need for human intervention.

During the exploration, we also need to map the gathered haptic data to the
corresponding position in the pre-segmented visual data obtained from the
previous step. This is done by tracking the position of the contact point
$\mathbf{x_c} \in \mathbb{R}^3$ through forward kinematics, and finding the
point $y$ in the object point cloud that has the smaller Euclidean distance to
the current contact point ($\mathbf{x_p} \approx \mathbf{x_c}$). We then assign
to $y$ the friction coefficient computed by \eqref{eq:eq_mu} at $\matr{x_c}$.
The benefit of using this approach is that some uncertainties caused by
calibration procedure can be reduced. After the exploration and tracking, the
explored points include both visual $\mathbf{V}$ and haptic data $\mathbf{H}$,
which can be represented as $y = (\mathbf{x_p},\mathbf{V},\mathbf{H})$. The next
step is to learn a model from the \expreg{} and use it to infer haptic feature
for \unexpreg{}.

\subsection{Variable friction model using visuo-haptic measurements}

We consider a visuo-haptic dataset $\xi = \{\xi_j\}_{j=1}^N$ defined by N
observations $\xi_j\in\mathbb{R}^D$. Each datapoint $\xi_j$ is represented with
input/output components indexed by $\mathbf{V}$ and $\mathbf{H}$, so that $\xi_j
= \genfrac[]{0pt}{1}{\xi_j^{\mathbf{V}}}{\xi_j^{\mathbf{H}}}$ and $D =
D^\mathbf{V} + D^\mathbf{H}$. In this work, $\xi_j$ is a concatenation of visual
features, (input component) and haptic features (output component). 

As the haptic exploration is conducted only once on a linear path along the
object, the tip of the robot only touches a few points in each \expregsing{}.
Thus, the number of points that has been assigned a friction coefficient value
is always smaller than the total number of points of each \expregsing{}.
Therefore, the covariance between RGB color components is computed using all of
the points in the region while the covariance between friction coefficient and
color components are computed using only the points that are touched by the
robot. Then, given the visuo-haptic dataset $\xi$, we use a \ac{gmm} with
$C$-components optimized through \ac{em} to encode the joint probability
distribution $\mathcal{P}(\xi^\mathcal{I},\xi^\mathcal{O})$. After a \ac{gmm} is
fitted to the dataset, \ac{gmr} can subsequently be used to estimate haptic
features $\xi_*^\mathcal{O}$  for visual features $\xi_*^\mathcal{I} \in
\mathbb{R}^D$ of \unexpreg{} as mentioned in \secref{sec:prob_form}.

Additionally, as discussed in \secref{sec:prob_form} we include a background
component in the \ac{gmm} model to reflect the estimation uncertainty. The
background component is constructed using measurements from the entire scene,
thus representing the entire variability. If an input, \ie{} visual feature, is
not close to any main components, the corresponding output will depend primarily
on the background component. This allows the model to capture the estimation
uncertainty such that a high variance is predicted when a particular region is
not visually similar to any of the regions with haptic measurements. This
uncertainty measurement can be used to actively make requests for new haptic
explorations in uncertain regions.

\section{{Experiments and Results}}
\label{sec:exp_and_res}

To demonstrate the ability of the proposed approach to estimate surface friction
without being restricted by the assumption that objects have uniform homogeneous
friction everywhere, we first evaluate the capabilities of the method by testing
it with different objects, including multi-material objects. Furthermore, we
report on the repeatability of the results. Afterwards, we present a grasping
case study in order to demonstrate the benefits of accurate friction estimation
in practical robotics applications.

\subsection{Experimental setup}

\begin{figure}
    \centering
    \includegraphics[width=.8\linewidth]{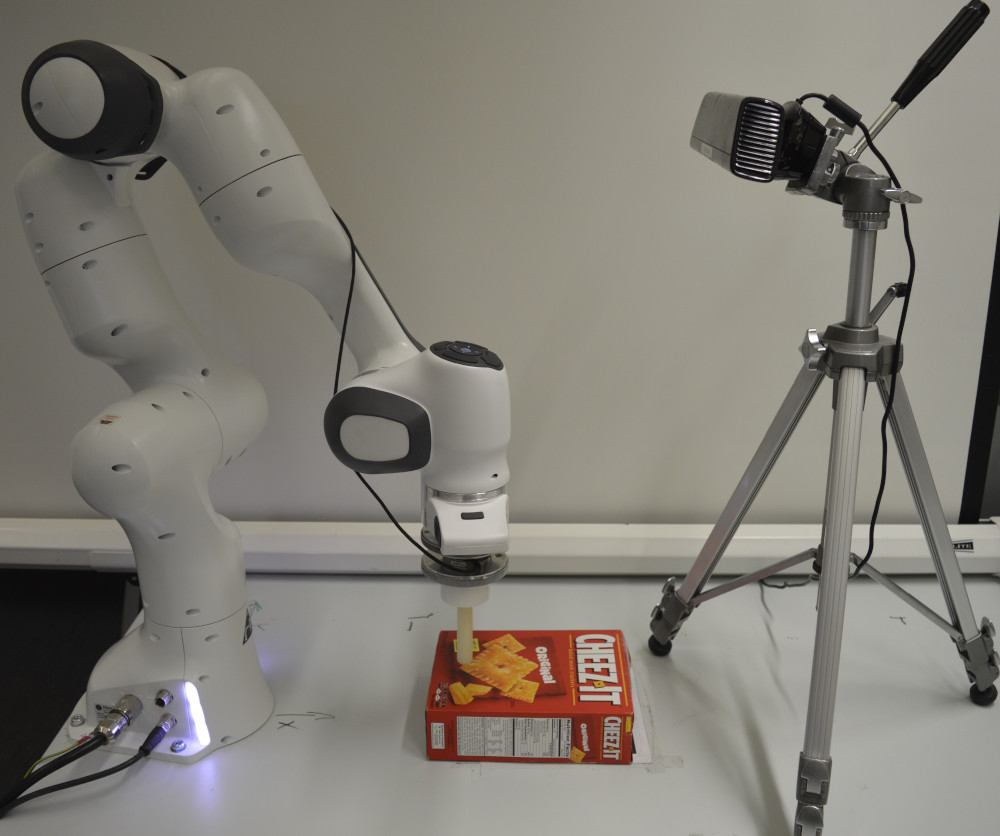}
    \caption{The experimental setup}
    \label{fig:setup}
    \vspace{-1.2em}
\end{figure}

The experiments are performed using the \panda{} robot and a Kinect
360\si{\degree} camera to capture the input \pc{}s as shown in
\figref{fig:setup}. We used an Aruco marker for the extrinsic calibration of the
camera. Once the input \pc{} was captured, it was filtered and pre-segmented as
explained in \secref{sec:implementation}. To perform the haptic exploration, we
used a six-axis force–torque sensor (ATI Mini45) attached between the robot's
wrist and the gripper. The haptic exploration is performed once for each object.

\subsection{Model representation with real robot and objects}

\begin{figure*}
    \centering
    \expfig{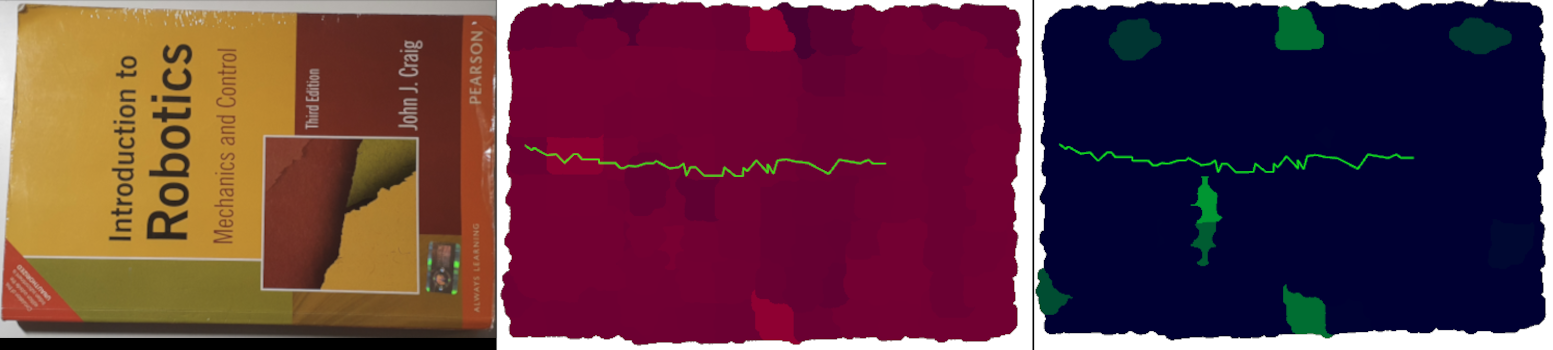}{Book}
    \expfig{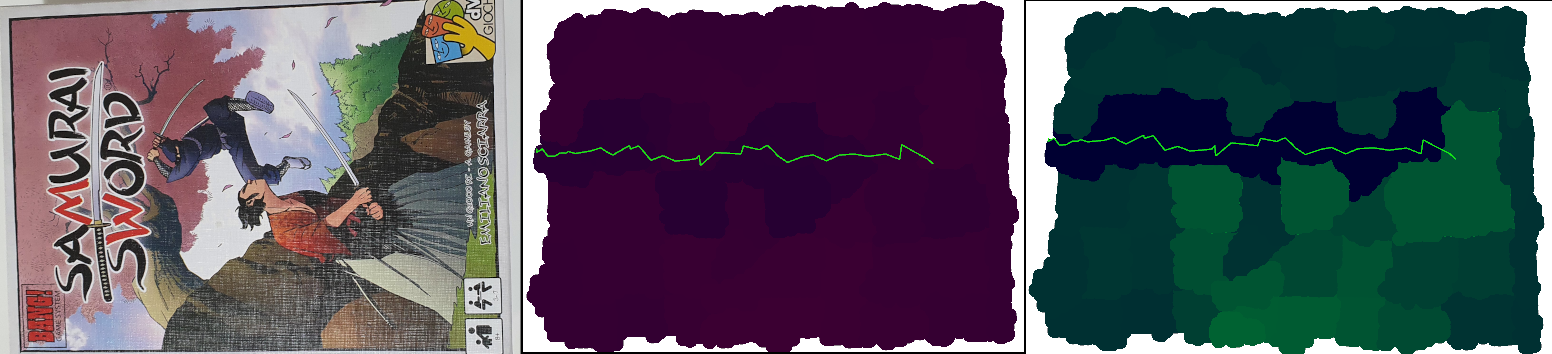}{Boardgame}
    \expfig{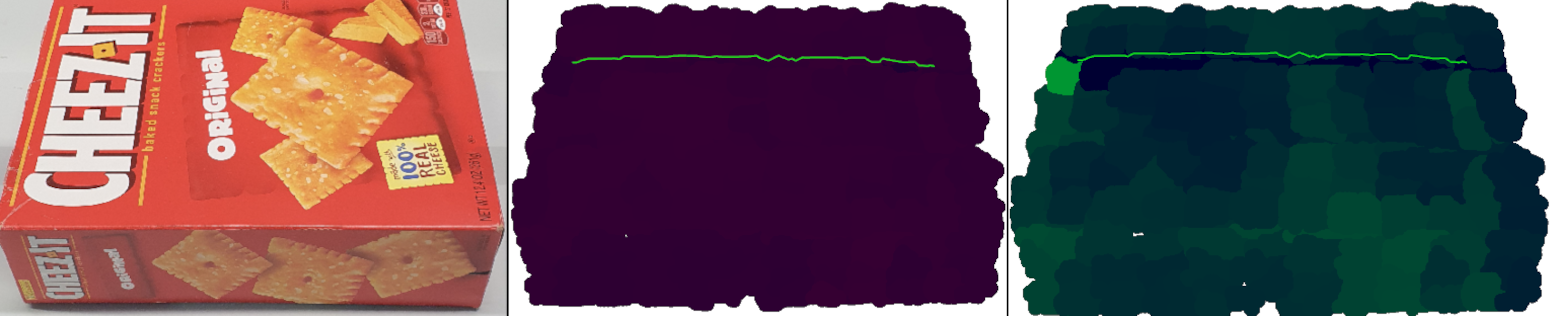}{Box}
    
    \expfig{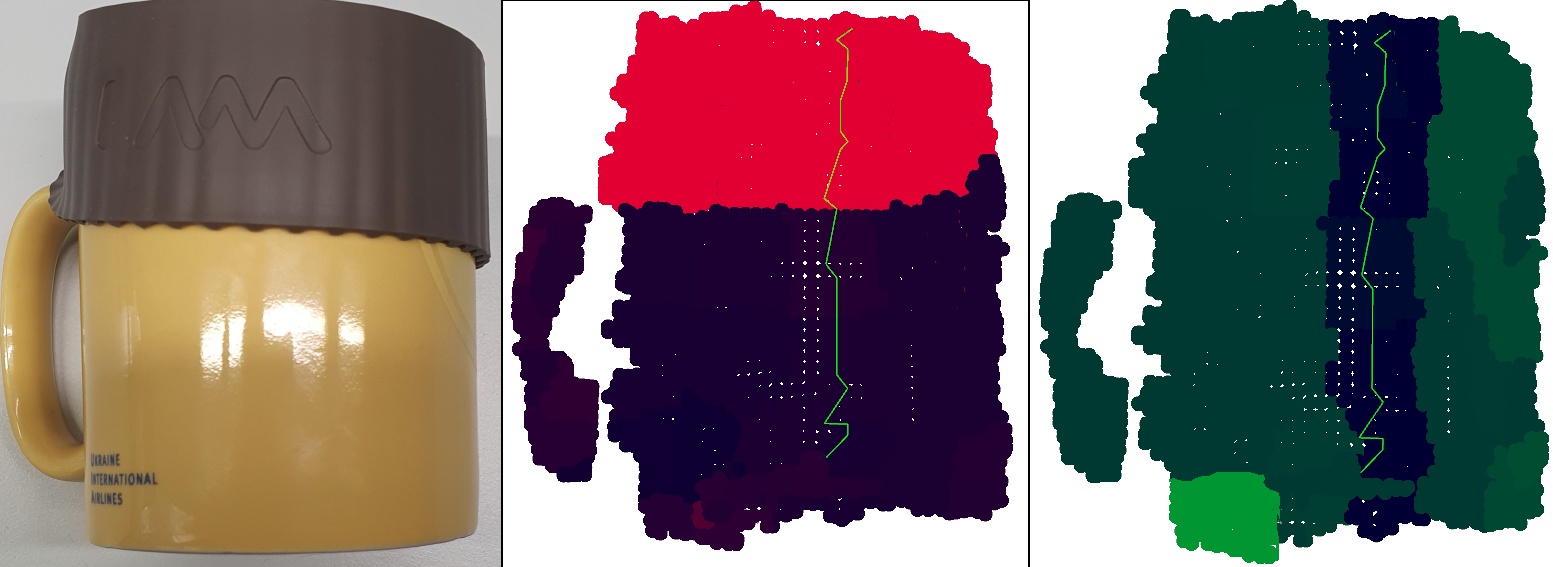}{Yellow mug}
    \expfig{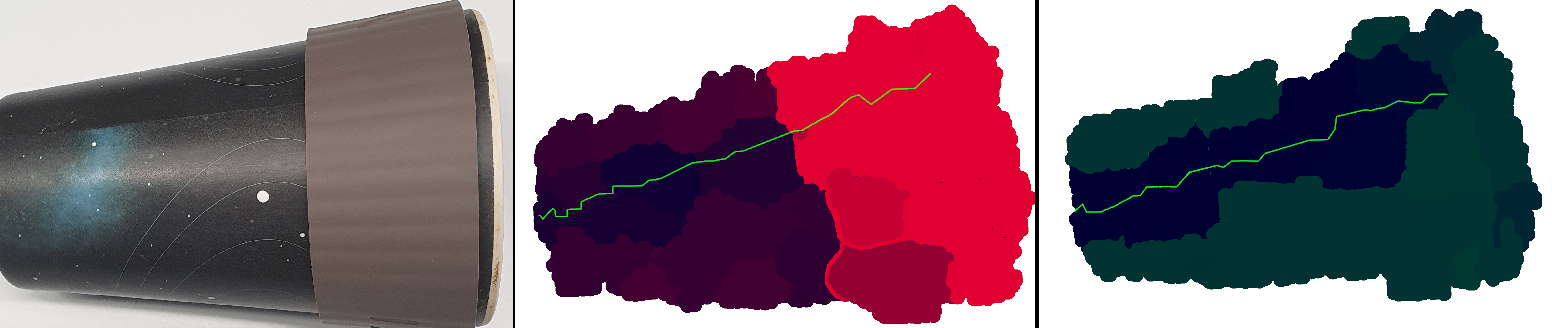}{Black cup}
    \expfig{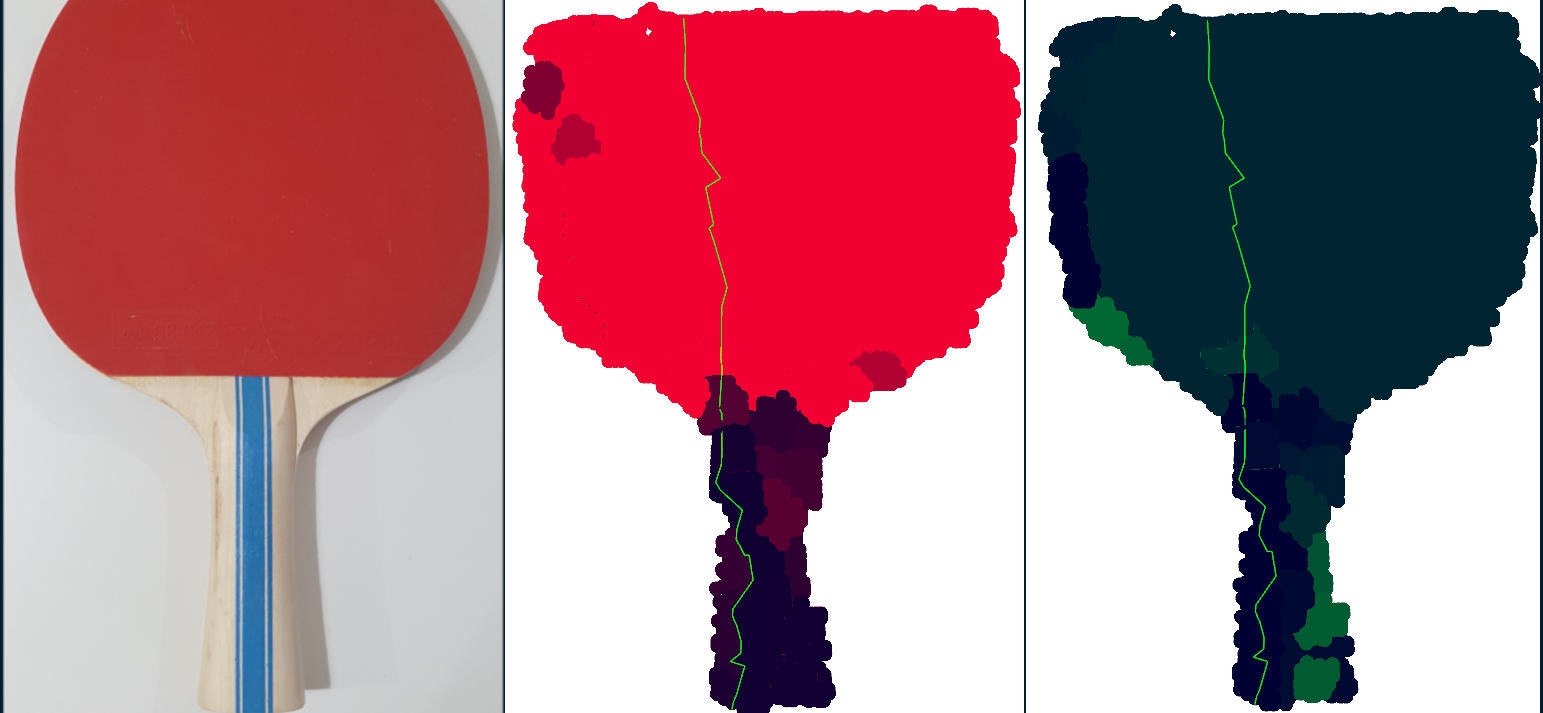}{Table tennis racket}
    
    \expfig{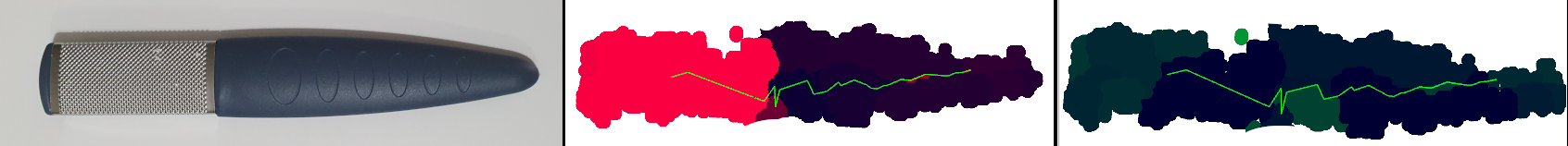}{Rasp}
    \expfig{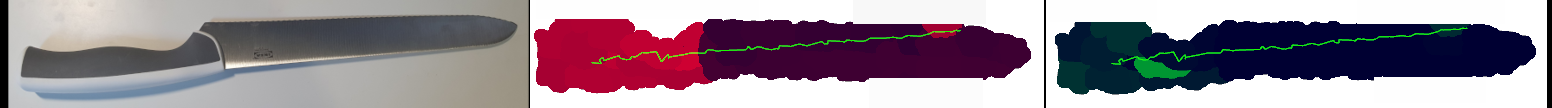}{Bread knife}
    \expfig{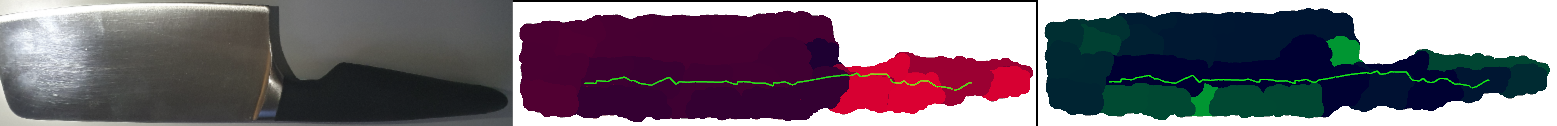}{Butcher knife}
    
    \expfig{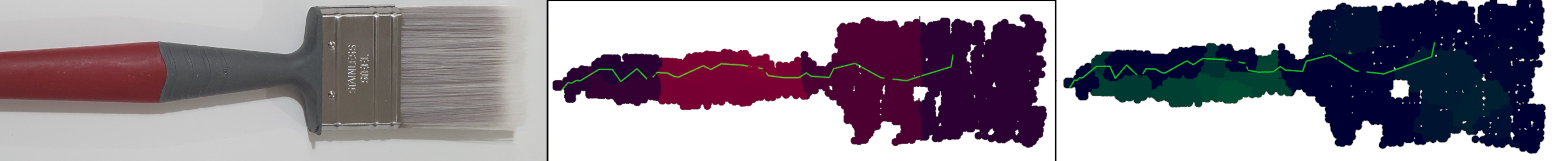}{Brush}
    \expfig{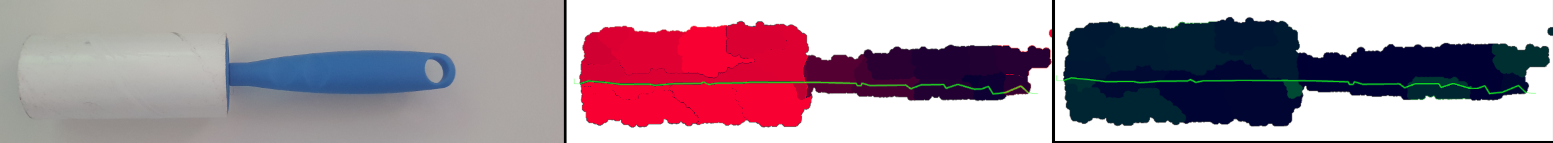}{Cloth roller}
    \expfig{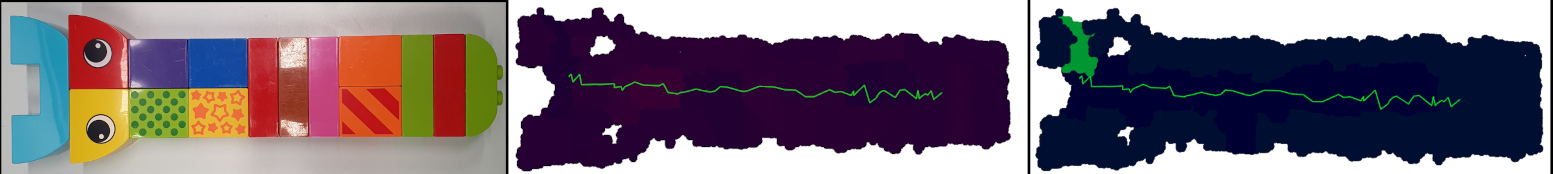}{Toy blocks}
    
    \expfig{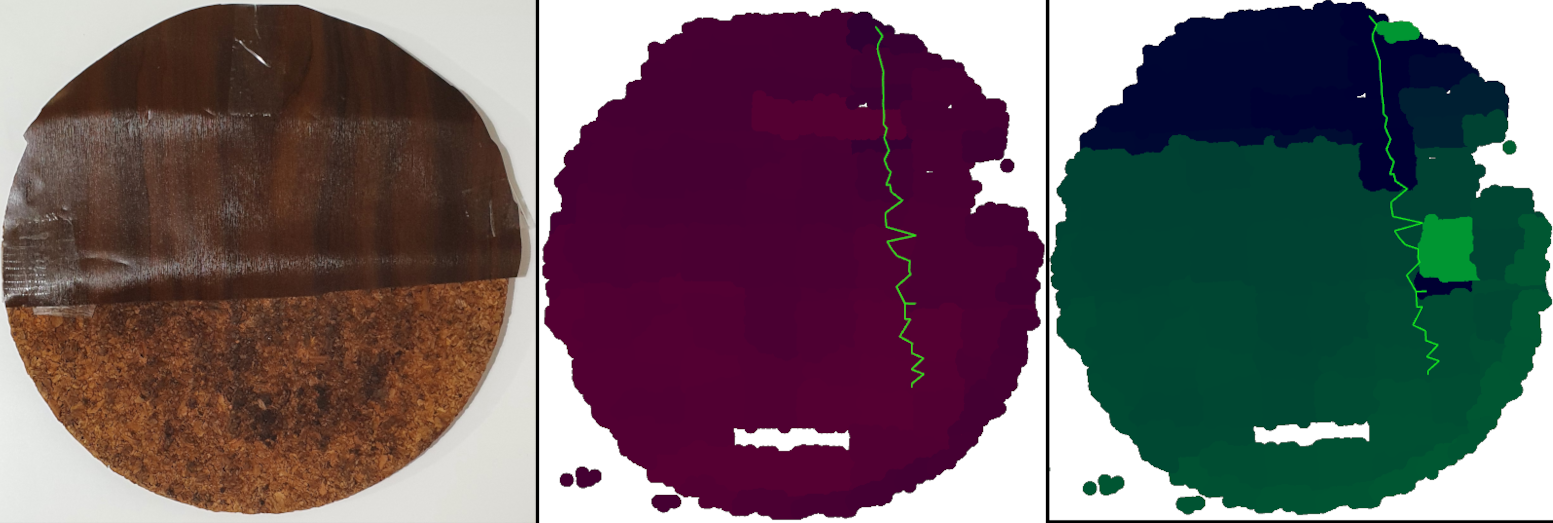}{Cork trivet}
    \expfig{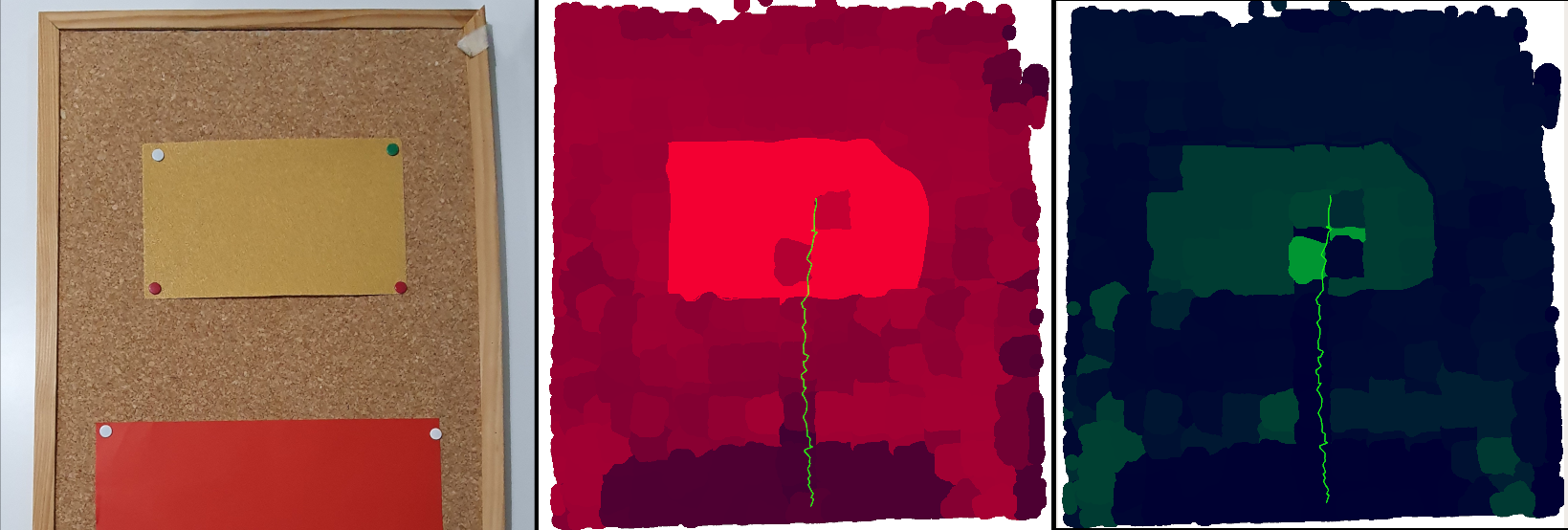}{Board (yellow:sandpaper, red:paper)}
    \expfig{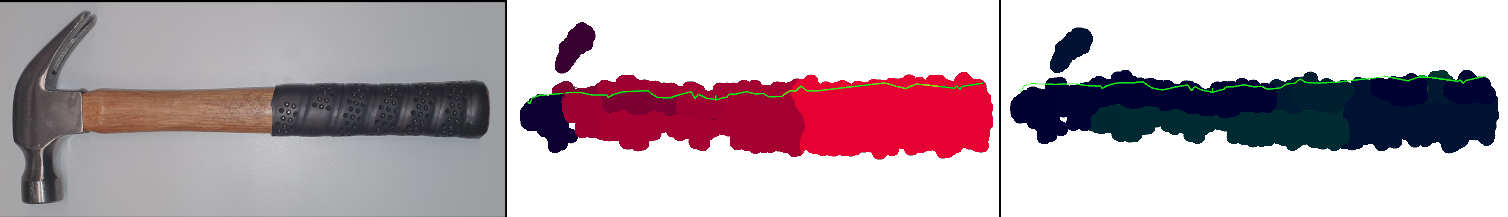}{Hammer}
    \caption{Real objects, along with estimated friction coefficient and
    uncertainty returned by the proposed method. Red indicates higher friction
    coefficient value, while green denotes higher uncertainty. The haptic
    exploration is showed by the path in green \bestcolor{}.}
    \label{fig:all_experiment}
    \vspace{-1em}
\end{figure*}

To study the capability of the proposed method, we ran the experiment on fifteen
different objects, shown in \figref{fig:all_experiment}. Of these objects, the
book (\figref{fig:book}), the boardgame (\figref{fig:samurai}), the cereal box
(\figref{fig:box}) and the toy blocks (\figref{fig:lego}) represent
single-material objects, where friction coefficient is identical across the
whole surface, while the rest of the objects are composed of multiple materials
(ranging from two to four).

\figref{fig:all_experiment} shows qualitative results obtained from the proposed
method applied on the target objects. We show in the result the original object,
the estimated friction, and the estimated uncertainty of each target object
respectively. The haptic exploration is visualized as the green path. For the
estimated friction, regions are colored according to their friction coefficient
value, \ie{} more red corresponds to higher friction. For the estimated
uncertainty, the level of uncertainty is represented by green color, such that
the greener the higher uncertainty. The results show that the proposed method
has successfully estimated the friction coefficient for all the target objects
even in presence of multiple materials. Specifically, the method is able to not
only produce similar friction coefficients across the object surface in the case
of single-material object (\ie{} book and cereal box), but also to provide
different friction coefficients for different parts of objects in the case of
multi-material objects. For example, in the case of the yellow mug
(\figref{fig:yellow_mug}) the method estimated higher friction coefficient value
for the rubber band, and provided an accurate boundary between the rubber part
and the ceramic part of the object. Furthermore, the proposed method is shown to
work well even for objects that are made from more than two materials such as
the hammer (\figref{fig:hammer}) and the brush (\figref{fig:brush_hori}). In the
case of the brush, the method was able to segment the object into four parts
representing four different materials. This predicted friction map is accurate
compared to the real object where the handle is made from both plastic and
rubber, and the head is made from steel and fabric.

Another interesting point from the result is that our method works even in cases
where visual features vary a lot but haptic features are identical across the
surface, like the toy blocks and the board game. As this scenario would likely
cause problems to methods estimating friction only from vision as discussed in
\secref{sec:related_work}, this experiment shows the benefit of combining both
visual and haptic feedback for friction estimation. Additionally, in the case of
the butcher knife (\figref{fig:butcher}) with rubber handle, without
encapsulating haptic feedback, one would hardly know if the handle has lower
(made of plastic) or higher (made of rubber) friction coefficient compared to
the steel blade. Since we conducted haptic exploration across the object, our
method is capable of estimating higher friction coefficient for the handle,
which is consistent with the ground truth object. However, objects having
similar visual features for surfaces having different friction has proven
challenging for our approach. This is the case for the modified cork trivet
(\figref{fig:cork_trivet}), where we covered half of the surface with a
wood-textured wallpaper. The trivet is now composed of two materials with
similar visual features. The result shows that our method provides a uniform
friction map over the whole object surface. However, due to the variations in
the haptic features along the explored path, the estimated uncertainty is shown
to be high in the lower area of the trivet. One of the potential solution to
overcome this problem is to conduct new  haptic explorations in uncertain
regions as discussed in \secref{sec:implementation}.

Furthermore, the experimental results also show the uncertainty map of the
proposed method. In the case of the book and the cereal box, the uncertainty is
extremely low because there are no abnormal visual features. In other words, all
visual features of unexplored regions are almost as varied as the ones of
explored regions. On the other hand, for the yellow mug case, the uncertainty is
high at the left edge of the mug where the blue pattern is located. This result
is suspected to be because the haptic exploration is carried out only in the
middle of the mug, where the color is always yellow or brown. As the blue color
is far from the explored visual feature, that region is classified by the
background class, which in turn produces a high uncertainty for the prediction.
A similar result is shown in the case of the board game. However, in all these
cases, even though the uncertainty is high, the estimated friction is still
similar to that of the rest of the regions of the same material.

\subsection{Repeatability of the results}

\begin{figure}
    \centering
    \includegraphics[width=0.8\linewidth]{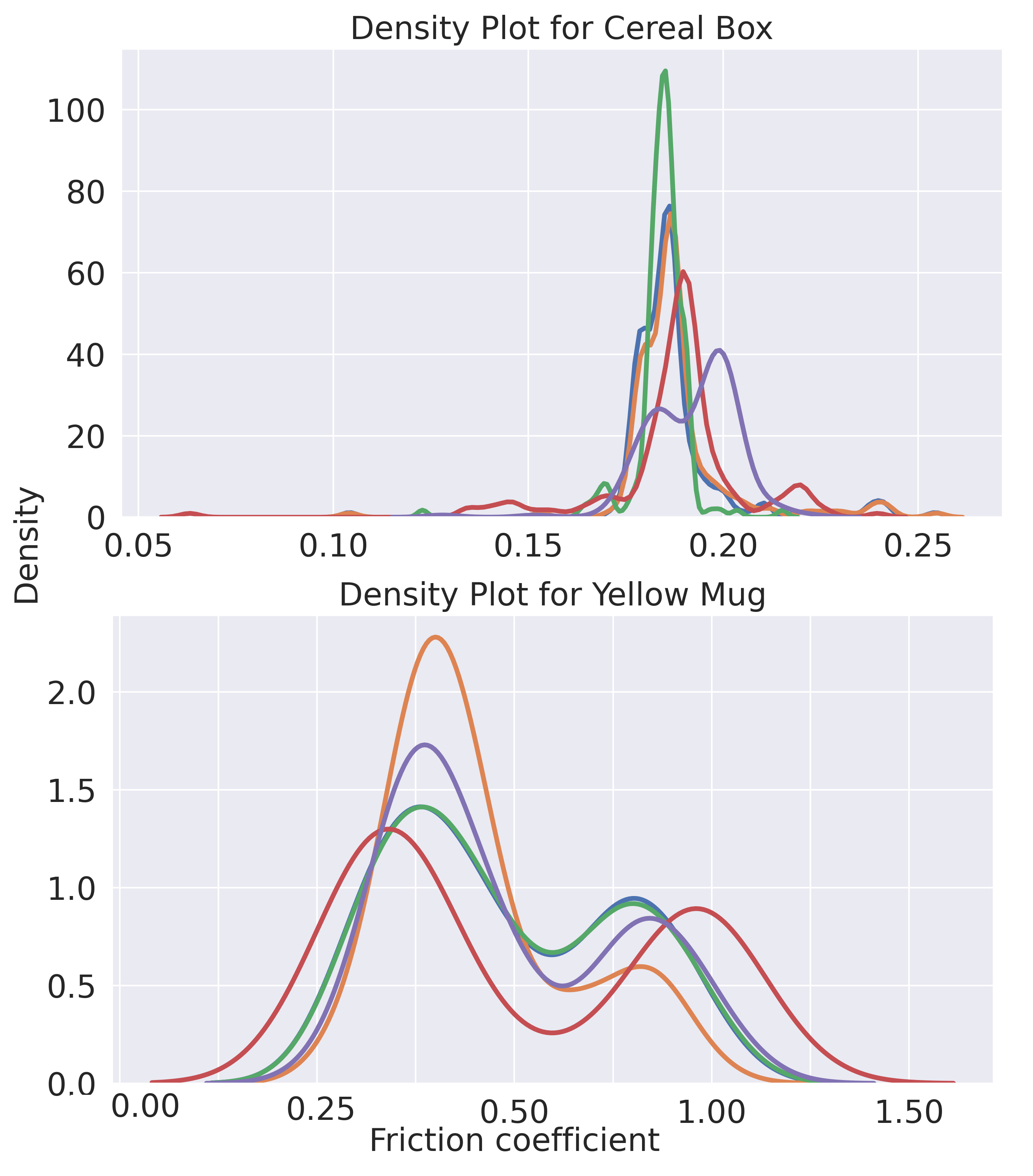}
    \caption{The density plots of the cereal box and the yellow mug, showing 
    consistent friction coefficient estimation over five repetitions of the 
    experiment.}
    \label{fig:repetability}
    \vspace{-1.2em}
\end{figure}

Next we evaluated the repeatability of the proposed method by running the
experiment five times on each object. The estimated friction coefficient is
recorded and plotted as density plots for analysis as shown in
\figref{fig:repetability}, where, for each friction coefficient, we show the
number of regions found having that value. For brevity, only the results of the
cereal box representing single-material objects and the yellow mug representing
multi-material objects are plotted for discussion. The results show that the
proposed method produces reasonable result in term of repeatability. Note that
despite small variation between repetitions, we still can clearly see that both
density plots represent accurately the number of the material in each case. For
example, in the case of the cereal box the density plot only has one peak which
denotes a single-material object while the density plot of the yellow mug
contains two peaks representing a multi-material object.  

\subsection{Grasping case study}

In robotic grasping, grasp selection is usually complex as its stability takes
into account different factors such as gripper geometry, object shape, mass, and
surface friction. Although methods have been proposed to grasp without needing
the estimation of physical properties \cite{adjible18}, most common approaches
for grasping rely on metrics based on physical properties to evaluate grasp
quality. In particular, a set of best grasps are sampled and evaluated from a
grasp sampling and evaluation algorithm. Typically, parameters like surface
friction are kept constant during the process; however, the selected grasps that
are executed in the real worlds may fail as well due to the contact with low
friction surface. In order to demonstrate the usefulness of our method, we
conducted a grasping case study where the estimated friction information is used
to sample and evaluate the grasps. 

In this demonstration, we first capture the target object from different
viewpoints and merge them together in order to obtain a multi-view \pc{} of the
object. The proposed method is then applied to the given \pc{} to produce the
surface friction estimation. As the input of the grasp sampler is typically a
mesh, we converted the estimated \pc{} to a mesh using Meshlab. Since the mesh
does not contain any information about the estimated friction, we calculate the
center of each face of the mesh, find its closet point and assign the friction
coefficient value of the point to the corresponded face. Next, the mesh with
assigned friction coefficient is fed to a grasp sampler to generate grasp
candidates. Grasp candidates are sampled using antipodal grasp sampling method.
Specifically, we randomly select a point on the mesh and assume this point as
the first contact point. At this first contact point, a direction ray that lies
inside of the friction cone is generated. If the ray intersects with a certain
face, the intersect point will be the second contact point, and the closing
vector is assured. Next, we randomly generate an approach vector along the
closing vector. The generated grasps will then be checked for different type of
collisions. The grasps that collide with the mesh are filtered out and the
grasps that are not in collision are then evaluated using Ferrari \& Canny L1
quality metric~\cite{ferraricanny}.

\begin{figure}
    \centering
    \begin{subfigure}{\linewidth}
		\centering
        \includegraphics[width=.4\linewidth]{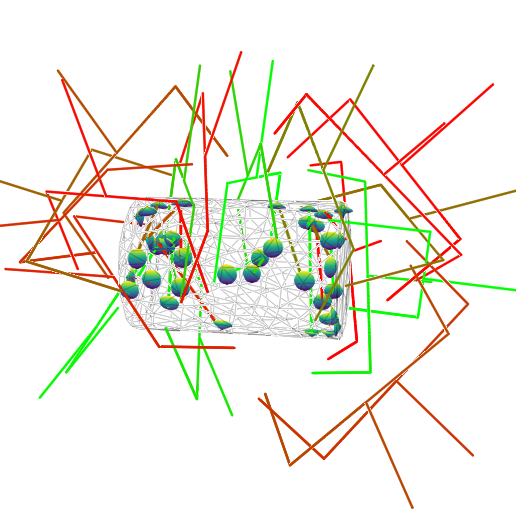}
		\includegraphics[width=.4\linewidth]{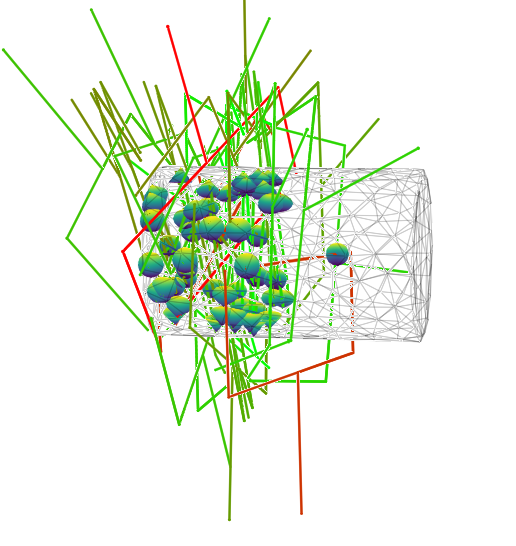}
	\end{subfigure}
    \begin{subfigure}{\linewidth}
		\centering
        \includegraphics[width=.4\linewidth]{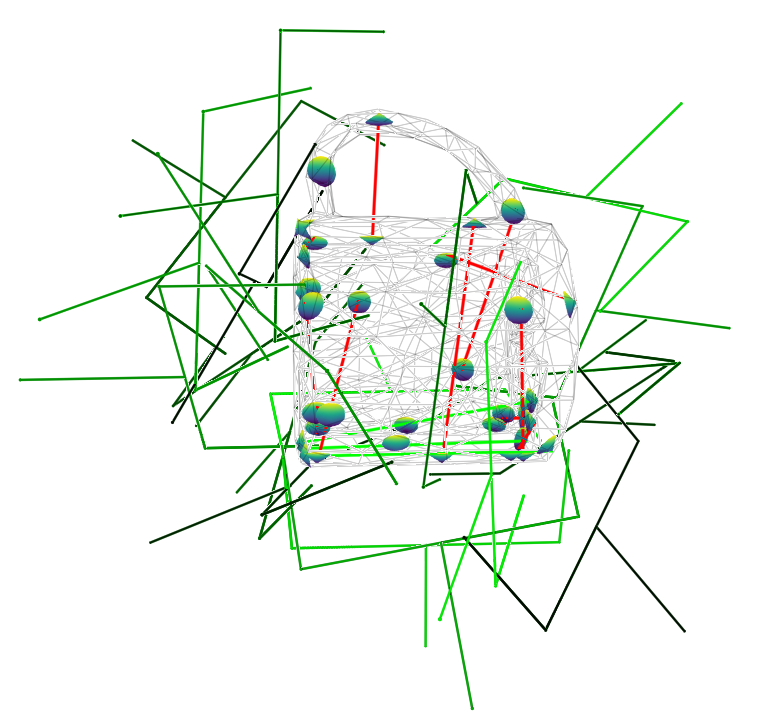}
		\includegraphics[width=.4\linewidth]{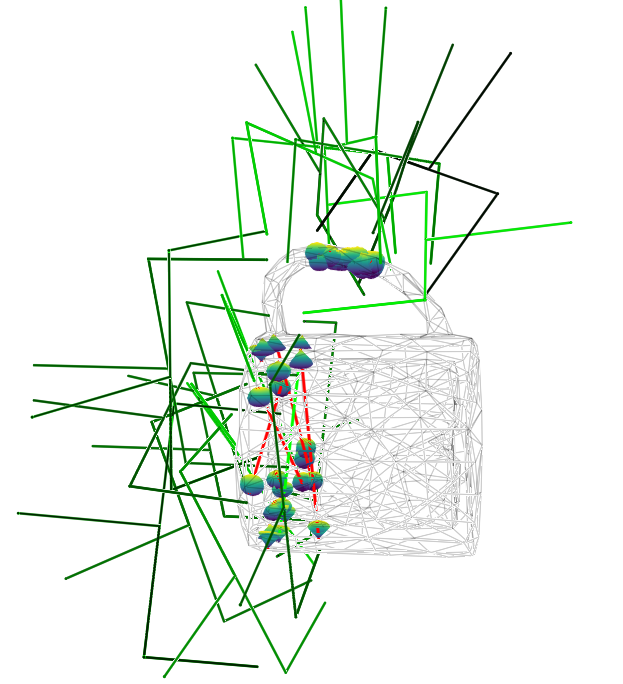}
	\end{subfigure}
    \caption{Grasp sampling results for two objects, black cup (top) and yellow
    mug (bottom), in two cases: using uniform friction (left), and non-uniform
    friction model (right).}
    \label{fig:sampler}
    \vspace{-1.5em}
\end{figure}

In this study, we sample 1000 grasps each on two object models: one with uniform
friction coefficient and one with non-uniform friction coefficient computed
using the proposed method. The coefficient of friction in the case of the
uniform friction was chosen similar to that of the high friction area computed
by our method. The object used in this study are the black cup and the yellow
mug as shown in \figref{fig:all_experiment}. Good grasp candidates together with
friction cones at contact points of both cases are presented in
\figref{fig:sampler}. These results show that the sampler behaves as expected:
In the case of the model with a uniform friction coefficient
(\figref{fig:sampler} left figures), the grasp candidates are distributed across
the entire object, while the grasp candidates on the non-uniform friction
coefficient model (\figref{fig:sampler} right figures) only appear around the
left side of the object, where the friction coefficient is high. 

\begin{figure}
    \centering
    \begin{subfigure}{\linewidth}
		\centering
        \includegraphics[width=.32\linewidth]{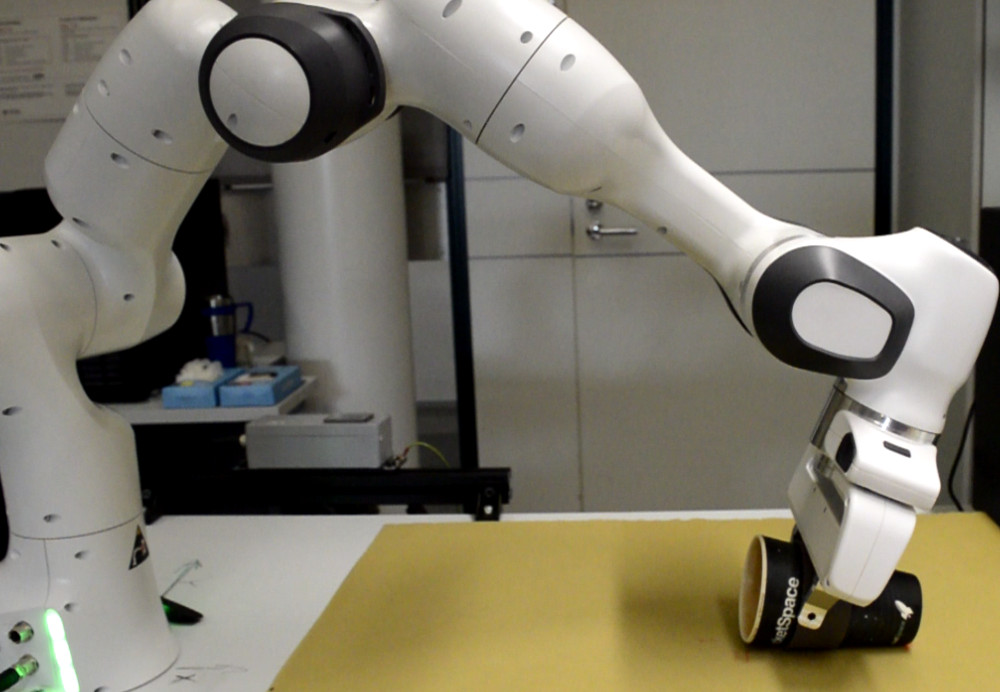}
		\includegraphics[width=.32\linewidth]{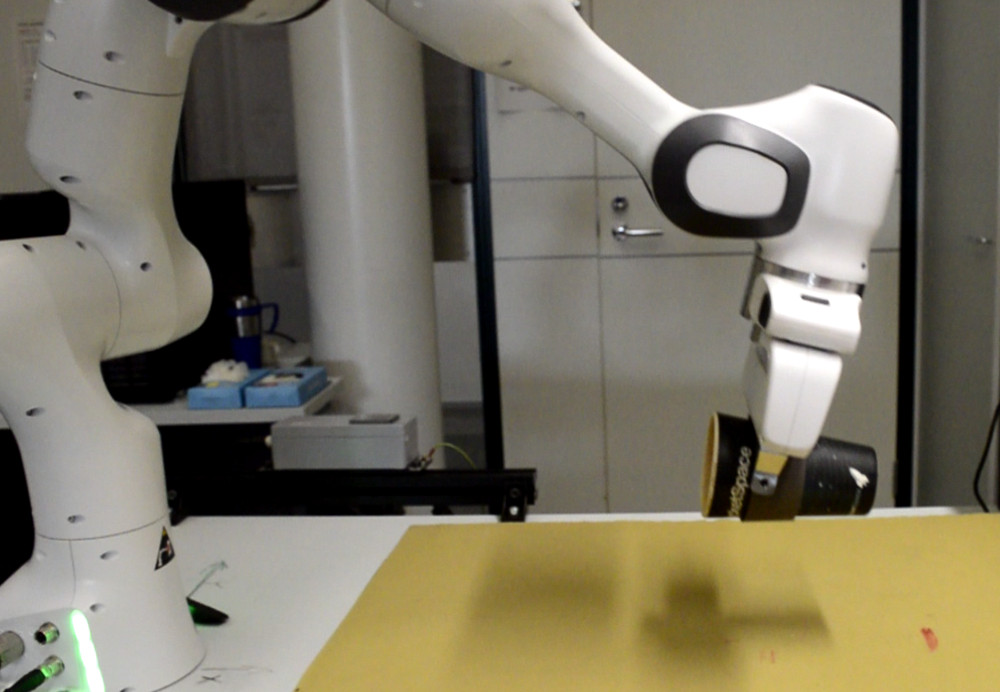}
	    \includegraphics[width=.32\linewidth]{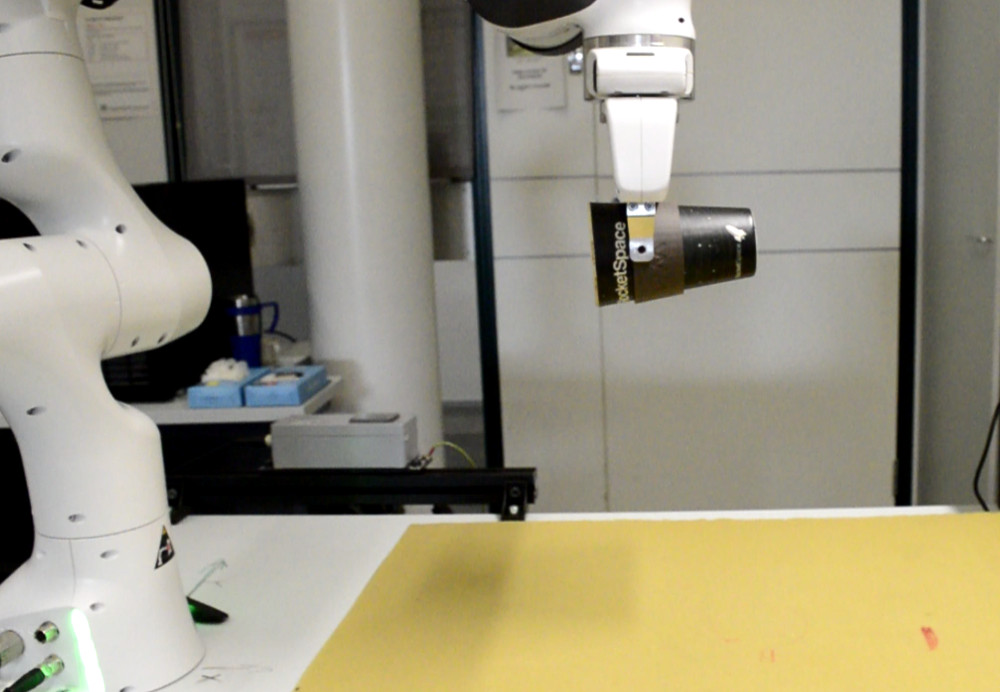}
	    \label{fig:grasping_1}
	    \vspace{0.5em}
	\end{subfigure}
	    \begin{subfigure}{\linewidth}
		\centering
        \includegraphics[width=.32\linewidth]{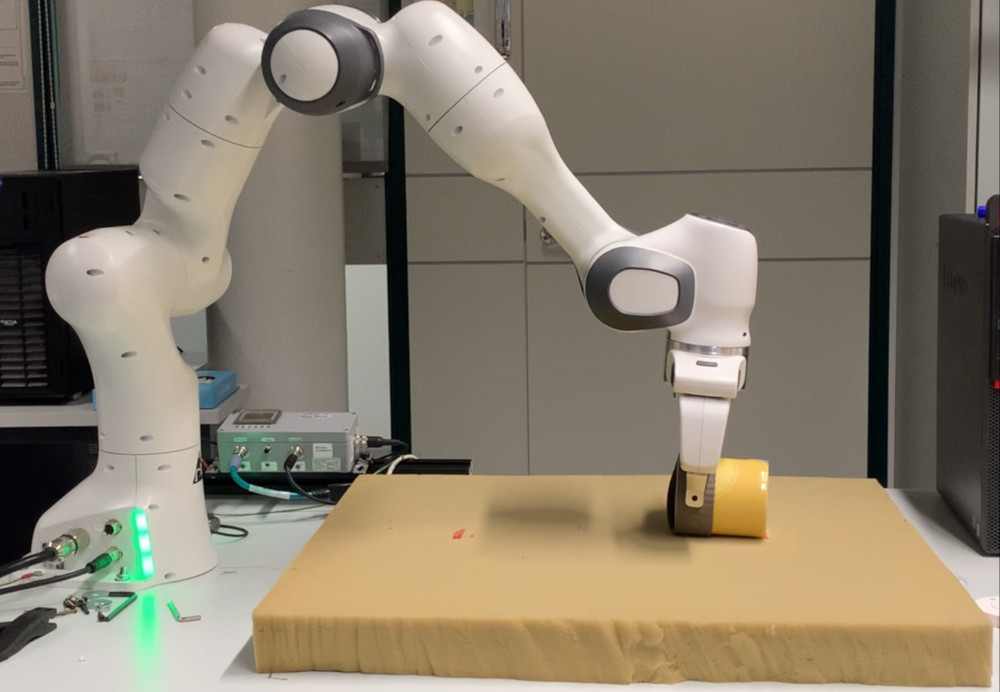}
		\includegraphics[width=.32\linewidth]{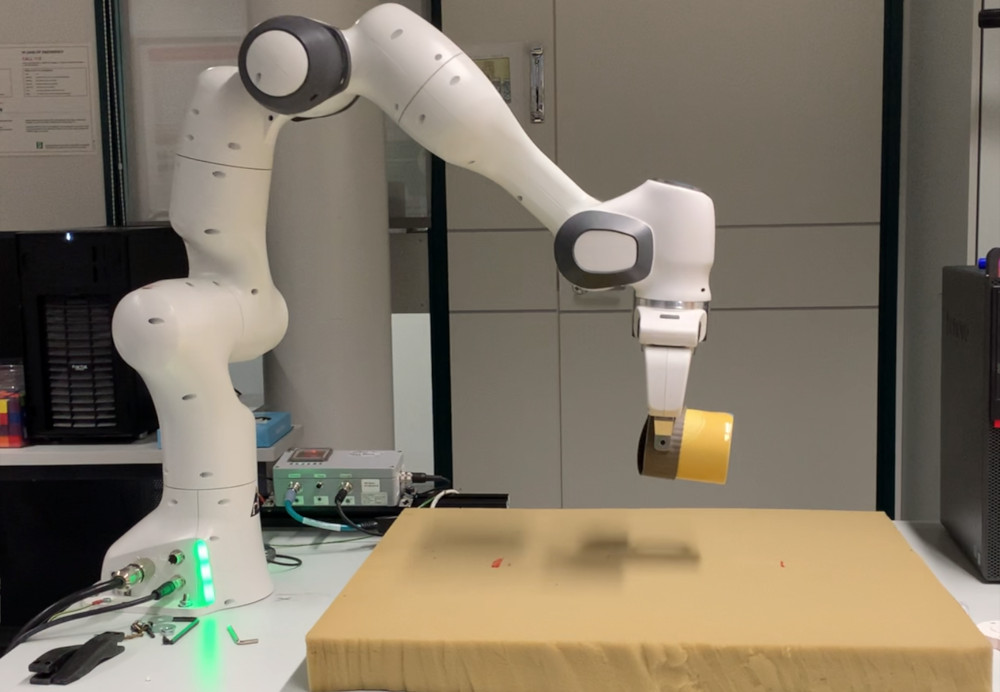}
	    \includegraphics[width=.32\linewidth]{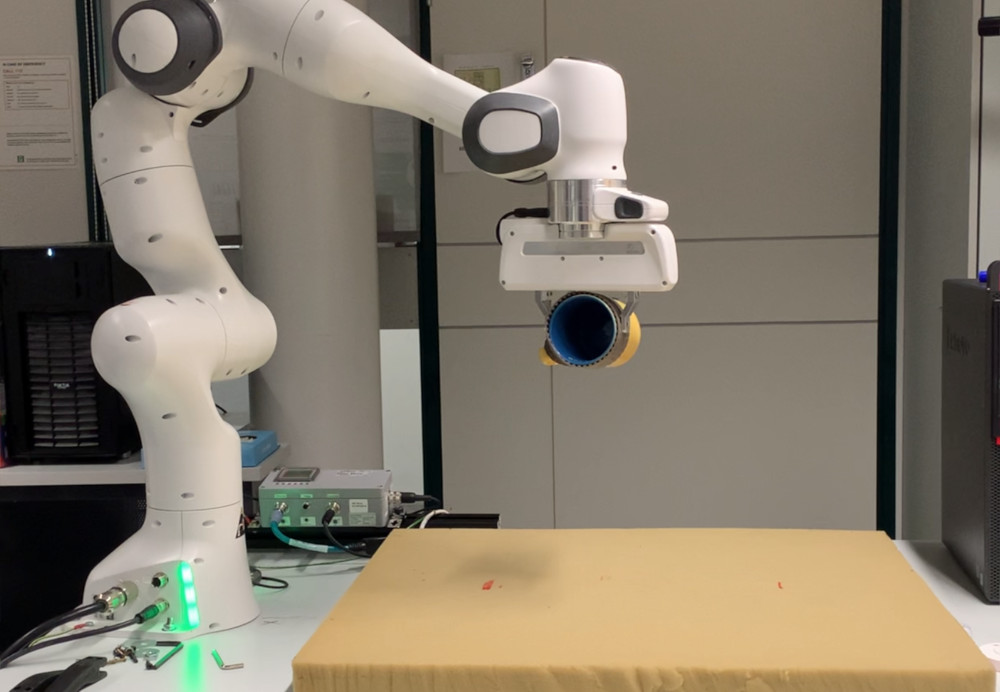}
	    \caption{Successful grasps executed on a region of high friction}
	    \label{fig:grasping_2}
	    \vspace{0.5em}
	\end{subfigure}
	    \begin{subfigure}{\linewidth}
		\centering
        \includegraphics[width=.32\linewidth]{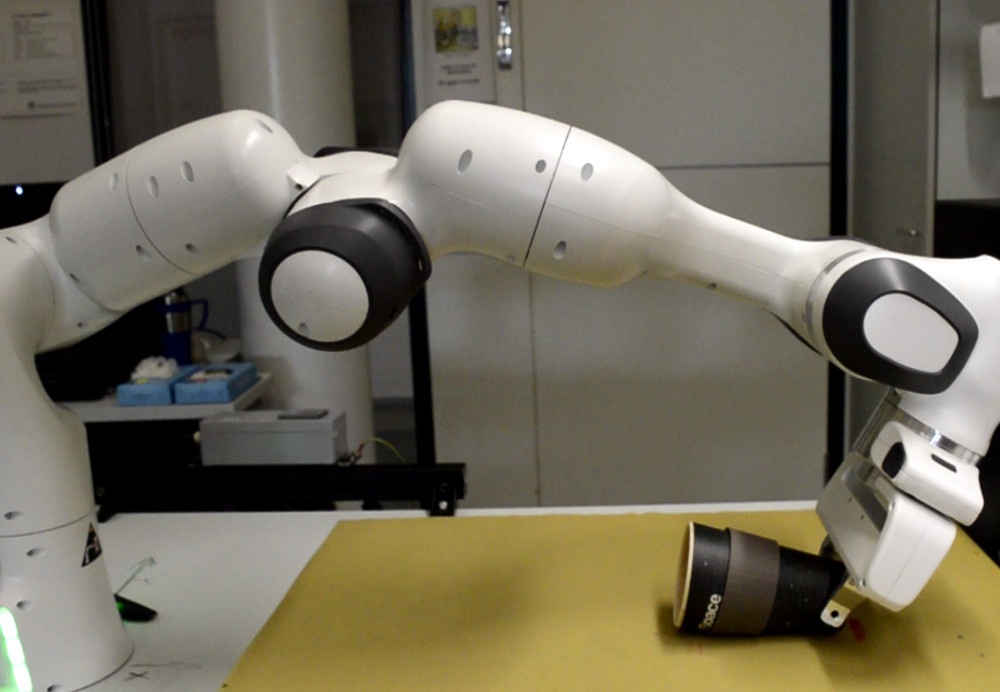}
		\includegraphics[width=.32\linewidth]{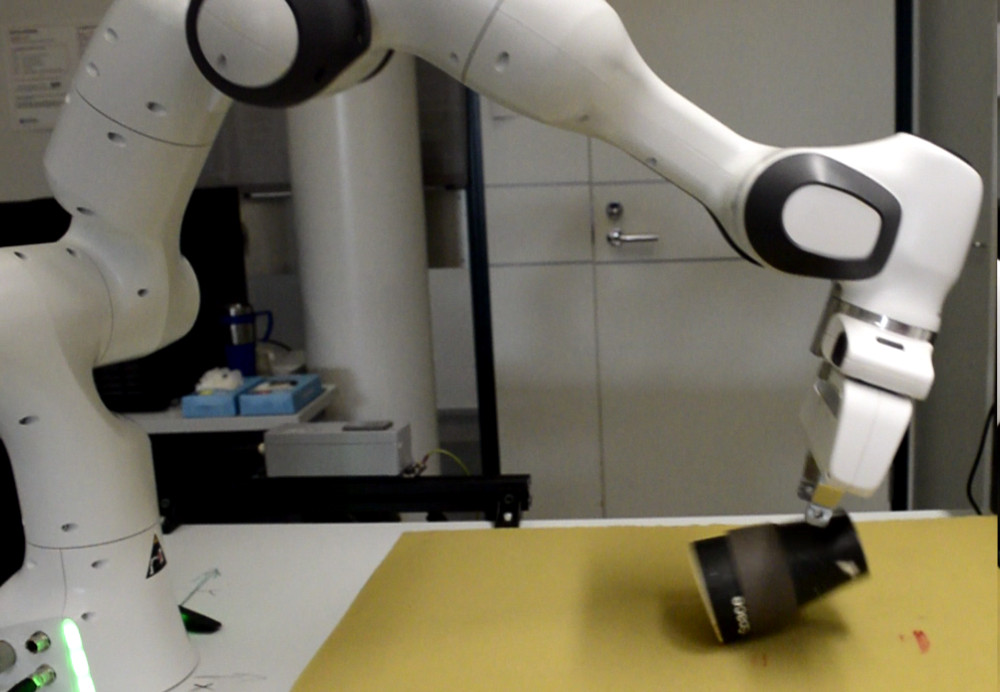}
	    \includegraphics[width=.32\linewidth]{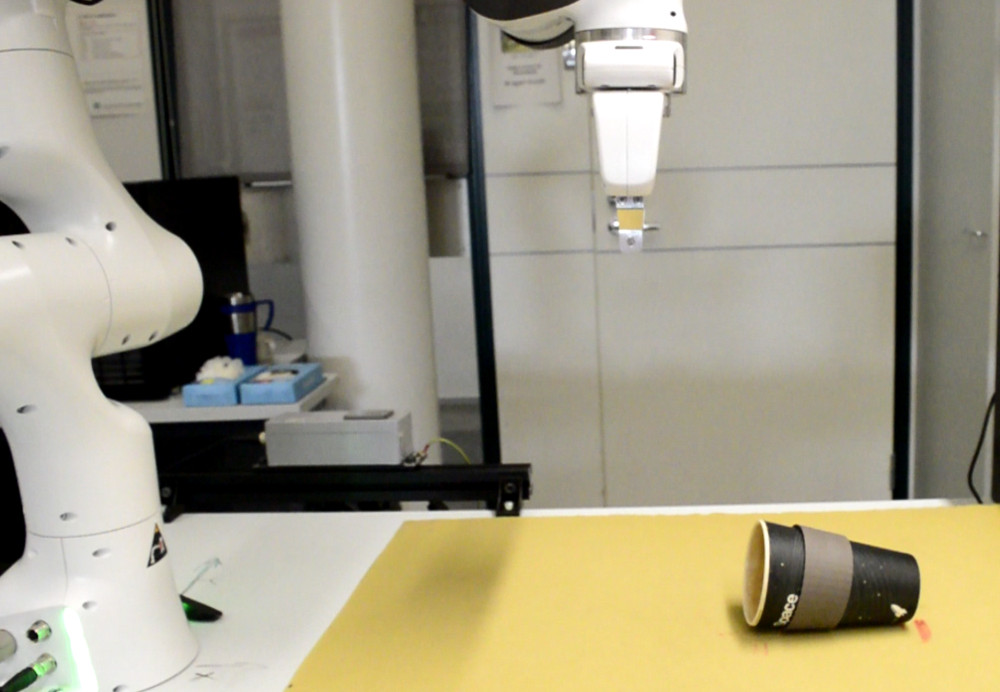}
	    \label{fig:grasping_3}
	    \vspace{0.5em}
	\end{subfigure}
    \begin{subfigure}{\linewidth}
		\centering
        \includegraphics[width=.32\linewidth]{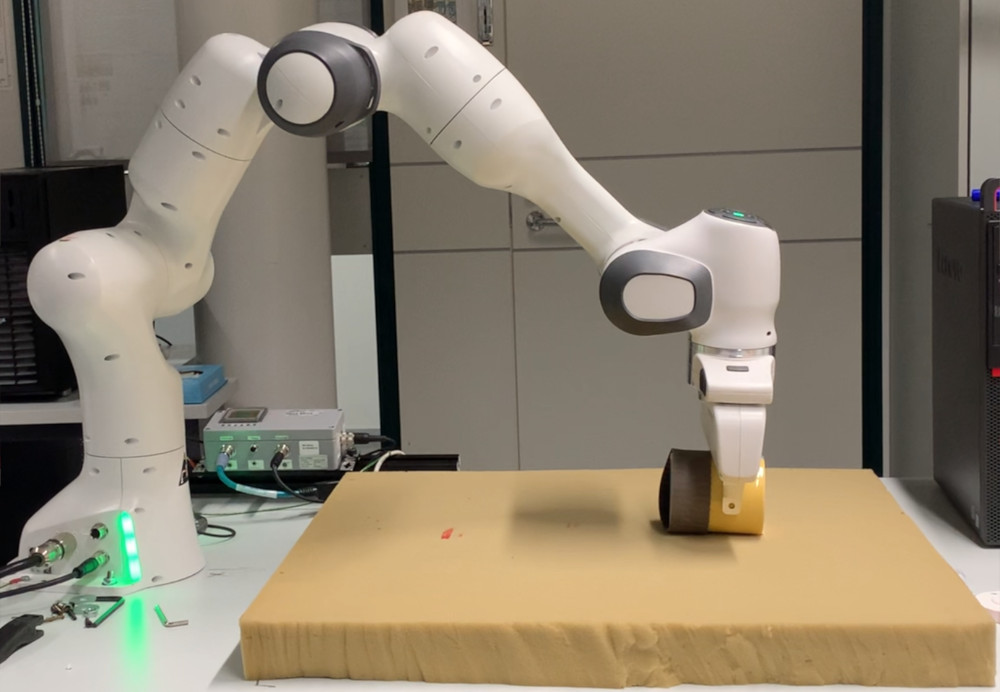}
		\includegraphics[width=.32\linewidth]{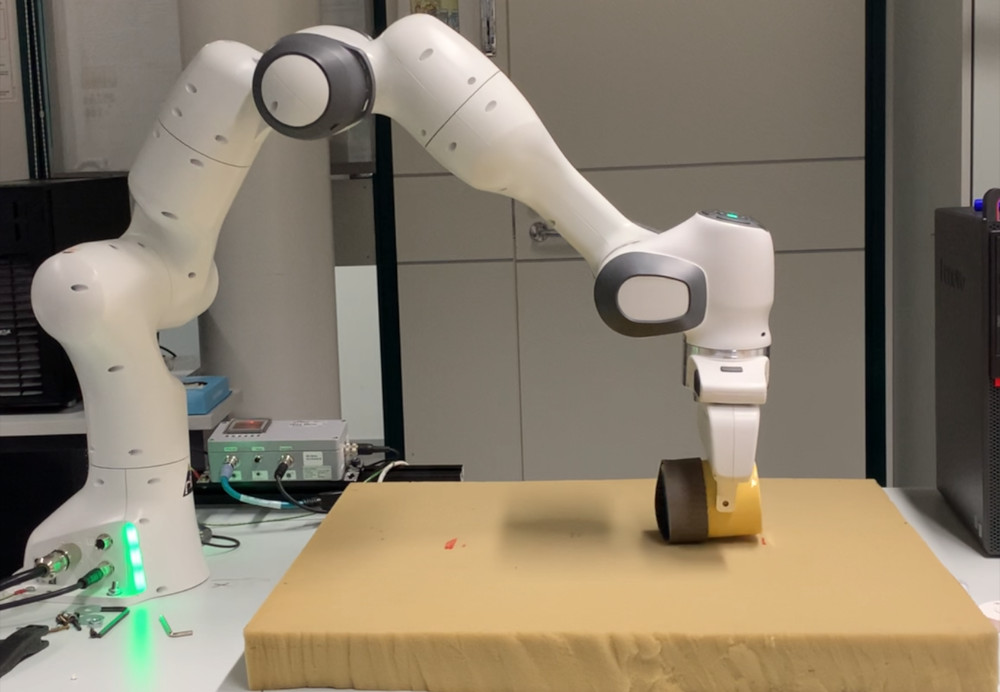}
	    \includegraphics[width=.32\linewidth]{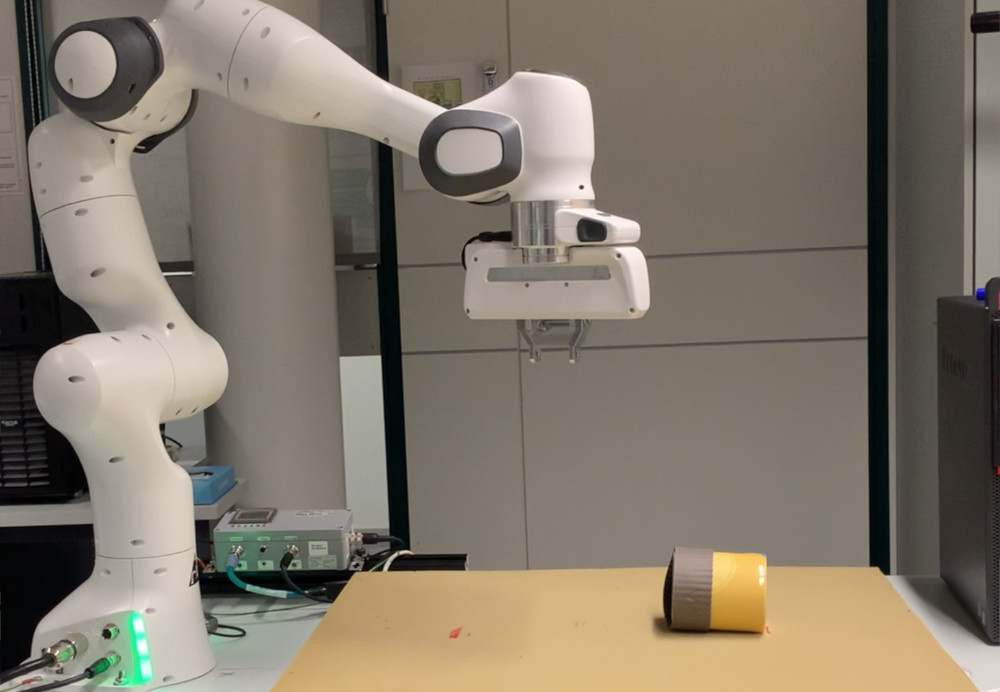}
	    \caption{Unsuccessful grasps executed on a region of low friction}
	    \label{fig:grasping_4}
	\end{subfigure}
    \caption{The robot executing different grasps on an object. The grasps are
    proposed by a sampler employing \subfigref{fig:grasping_2} the proposed
    friction model, or \subfigref{fig:grasping_4} a uniform friction model.}
    \label{fig:grasping}
    \vspace{-1.5em}
\end{figure}

To study the effect of the proposed method on grasp performance in the real
world, grasp candidates were generated using the antipodal grasp sampling
method. Five grasps with the highest quality according to the Ferrari-Canny
minimum quality metric were chosen in each case and executed with the real
robot. For the uniform friction coefficient case, we chose the five best grasps
that are on the right side of the object. By doing this, we can see how the
grasps actually behave if they make contact with low friction area of the
object. The grasps in both cases were executed with the same grasping force of
20 N. To evaluate if a grasp was successful, the robot moved to the planned
grasp pose, closed its fingers, and moved the arm back to the starting position.
Once there, the gripper was rotated around the last joint
(\figref{fig:grasping}). A grasp was consider successful if the object was
grasped in a stable manner for the whole procedure and unsuccessful if the
object was dropped \footnote{A video showcasing the grasping case study can be
found at\\ \url{https://irobotics.aalto.fi/video-friction/}}.

All evaluated grasp candidates generated with the sampler that utilized the
proposed method were successful as they grasped the object at the high friction
area. The chosen grasp candidates generated in the uniform friction case failed
to grasp the object since they aimed for the low friction area of the object,
which in turn, produced slippage during grasping. The results show that even
when the grasp sampler generates grasps with high quality under the uniform
friction case, the grasps still may fail when executed in the real world due to
incorrect friction assumption.

\section{Conclusions and future work}
\label{sec:conclusions}

We presented an approach that enables the estimation of local object physical
properties, like the surface friction coefficient, from visual and haptic cues,
which goes beyond the state-of-the-art by lifting the assumption that the target
object has uniform friction across its surface. The key component in this work
is the use of a probabilistic model to estimate the surface friction coefficient
of the unexplored areas from visuo-haptic data gathered by haptic exploration.
Furthermore, we also presented an approach to represent a level of uncertainty
of the estimate. This could be useful in future work to actively make requests
for new haptic explorations in the regions with high uncertainty. We
demonstrated the capability and the repeatability of the approach through
experiments on a wide range of objects including single-material and
multi-material objects. The results show that the proposed approach is capable
of providing object representations with varying surface friction coefficient.
Moreover, the friction coefficients can be used to guide grasp planning towards
areas of high friction, improving robotic grasping success rate. Despite the
good results, there is still room for improvements. For example, using more
robust and complex visual features, like the one that could be obtained by
modern deep learning approaches, could improve the quality and the robustness of
the estimated results against, \eg{} reflection and variations of lighting.


\section*{Acknowledgements}

We would like to thank Jens Lundell for the invaluable help in developing the
grasping experiments.

\bibliographystyle{IEEEtran}
\bibliography{refs}

\end{document}